\def\year{2020}\relax
\documentclass[letterpaper]{article} 
\usepackage{aaai20_arxiv}  
\usepackage{times}  
\usepackage{helvet} 
\usepackage{courier}  
\usepackage[hyphens]{url}  
\usepackage{graphicx} 
\usepackage{graphicx}  
\usepackage{bm}
\usepackage[ruled,vlined]{algorithm2e}
\usepackage{amsmath}
\usepackage{amsfonts}       
\usepackage{import}

\DeclareMathOperator*{\KL}{KL}
\DeclareMathOperator*{\KD}{KD}
\DeclareMathOperator*{\ARD}{ARD}
\DeclareMathOperator*{\argmax}{arg\,max}
\newcommand*\samethanks[1][\value{footnote}]{\footnotemark[#1]}

\title{Adversarially Robust Distillation}

\author{Micah Goldblum,\thanks{Authors contributed equally.} Liam Fowl,\samethanks[1] Soheil Feizi, Tom Goldstein\\ 
University of Maryland\\ 
4176 Campus Drive,\\
College Park, Maryland 20742\\
goldblum@umd.edu 
}

\begin{document}

\maketitle

\begin{abstract}
Knowledge distillation is effective for producing small, high-performance neural networks for classification, but these small networks are vulnerable to adversarial attacks.  This paper studies how adversarial robustness transfers from teacher to student during knowledge distillation.  We find that a large amount of robustness may be inherited by the student even when distilled on only clean images.  Second, we introduce Adversarially Robust Distillation (ARD) for distilling robustness onto student networks. In addition to producing small models with high test accuracy like conventional distillation, ARD also passes the superior robustness of large networks onto the student.  In our experiments, we find that ARD student models decisively outperform adversarially trained networks of identical architecture in terms of robust accuracy, surpassing state-of-the-art methods on standard robustness benchmarks.  Finally, we adapt recent fast adversarial training methods to ARD for accelerated robust distillation.
\end{abstract}

\section{Introduction}
\label{Introduction}

\par State-of-the-art deep neural networks for many computer vision tasks have tens of millions of parameters, hundreds of layers, and require billions of operations per inference \cite{He_2016,szegedy2016rethinking}.  However, networks are often deployed on mobile devices with limited compute and power budgets or on web servers without GPUs.  Such applications require efficient networks with light-weight inference costs \cite{dziugaite2016study,Sandler_2018,BNN,tai2015convolutional}.

\par \emph{Knowledge distillation} was introduced as a way to transfer the knowledge of a large pre-trained teacher network to a smaller light-weight student network \cite{HintonDistillation}. Instead of training the student network on one-hot class labels, distillation involves training the student network to emulate the outputs of the teacher. Knowledge distillation yields compact student networks that surpass the performance achievable by training from scratch without a teacher \cite{HintonDistillation}.

\par Classical distillation methods achieve high efficiency and accuracy but neglect security. Standard neural networks are easily fooled by \textit{adversarial examples}, in which small perturbations to inputs cause mis-classification \cite{FGSM,szegedy2013intriguing}. This phenomenon leads to major security vulnerabilities for high-stakes applications like self-driving cars, medical diagnosis, and copyright control \cite{santana2016learning,medical,saadatpanah2019adversarial}.  In such domains, efficiency and accuracy are not enough -- networks must also be adversarially robust. 

\par
\begin{figure*}[h!]
    \centering
    \includegraphics[scale = .55]{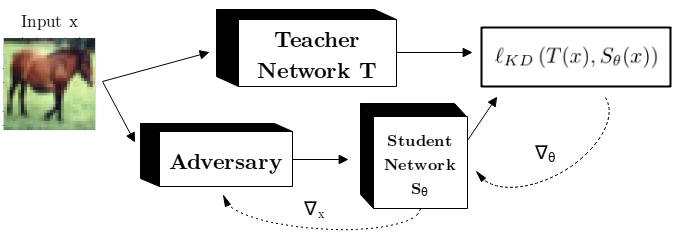}
    \caption{\small Adversarially Robust Distillation (ARD) works by minimizing discrepancies between the outputs of a teacher on natural images and the outputs of a student on adversarial images.  }
    \label{fig:ARD_diagram}
\end{figure*}

We study distillation methods that produce robust student networks.  Unlike conventional adversarial training \cite{Madry1,AdversarialTraining}, which encourages a network to output correct labels within an $\epsilon$-ball of training samples, the proposed \emph{Adversarially Robust Distillation} (ARD) instead encourages student networks to mimic their teacher's output within an $\epsilon$-ball of training samples (See Figure \ref{fig:ARD_diagram}).  Thus, ARD is a natural analogue of adversarial training but in the context of distillation.  Formally, we solve the minimax problem
\begin{equation}\label{eq:ARD}
\begin{aligned}
\min_{\theta}\mathbb{E}_{(X,y)\sim\mathcal{D}}\Bigg[\alpha t^2 \underbrace{\KL(S_{\theta}^t(X+\delta_{\theta}),T^t(X))}_{\text{Adversarially Robust Distillation loss}}\\ + (1-\alpha)\underbrace{\ell(S_{\theta}^t(X), y)}_{\text{classification loss}} \Bigg], \end{aligned}
\end{equation}
where $\delta_\theta =  \argmax_{\|\delta\|_{p}<\epsilon}\ell(S_{\theta}^t(X+\delta), y)$, $T$ and $S$ are teacher and student networks, and $\mathcal{D}$ is the data generating distribution.  See Section \ref{ARD} for a more thorough description.  Below, we summarize our contributions in this paper: 

\begin{itemize}
    \item We show that knowledge distillation using only natural images can preserve much of the teacher's robustness to adversarial attacks (see Table \ref{table:robustnessTransferred}), enabling the production of efficient robust models without the expensive cost of adversarial training.
    \item We introduce Adversarially Robust Distillation (ARD) for producing small robust student networks.  In our experiments, ARD students exhibit higher robust accuracy than adversarially trained models with identical architecture, and ARD often exhibits higher natural accuracy simultaneously.  Interestingly, ARD students may exhibit even higher robust accuracy than their teacher (see Table \ref{table:robustnessTransferred}).
    \item We accelerate our method for efficient training by adapting fast adversarial training methods.
\end{itemize}

\begin{figure*}[t!]
    \centering
    \includegraphics[scale = .6]{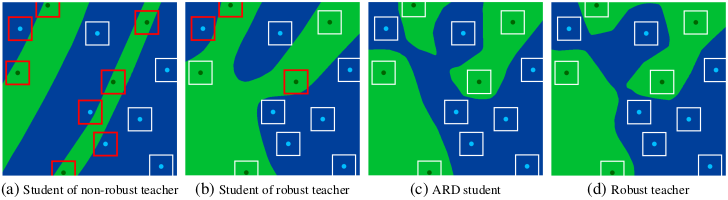}
\caption{\small{A student distilled from a robust teacher is more robust than a naturally trained network, but ARD produces a more robust network than either and closely mimics the teacher's decision boundary.  Adversarially vulnerable training points have $\ell_\infty$ boxes outlined in red.}}
\label{fig:DecisionBoundaries}
\end{figure*}

\begin{table}[h!]
\begin{center}
\begin{tabular}{|l|l|l|}
\hline
Model                           & \shortstack{Robust Accuracy \\ ($\mathcal{A}_{adv}$)} \\ \hline
AT ResNet18 teacher            & 44.46\% \\ \hline
AT ResNet18 $\to$ MobileNetV2                  & 38.21\%  \\ \hline
AT ResNet18 $\xrightarrow{\text{\textbf{ARD}}}$ MobileNetV2                  & \textbf{50.22}\%  \\ \hline
\end{tabular}
\end{center}
\caption{Performance of an adv. trained (AT) teacher network and its student on CIFAR-10, where robust accuracy ($\mathcal{A}_{adv}$) is with respect to a $20$-step PGD attack as in \cite{Madry1}.  ``$\to$'' denotes ``knowledge distillation onto''. ``$\xrightarrow{\text{\textbf{ARD}}}$'' denotes ``adversarially robust distillation onto''.}
\label{table:robustnessTransferred}
\end{table}
Table \ref{table:robustnessTransferred} shows that a student may learn robust behavior from a robust teacher, even if it only sees clean images during distillation. Our method, ARD, outperforms knowledge distillation for producing robust students. To gain initial intuition for the differences between these methods, we visualize decision boundaries for a toy problem in Figure \ref{fig:DecisionBoundaries}. Randomly generated training data are depicted as colored dots with boxes showing the desired $\ell_\infty$ robustness radius. Background colors represent the classification regions of the networks (10-layer teacher and 5-layer student). In each case, the network achieves perfect training accuracy.  A training point is vulnerable to attack if its surrounding box contains multiple colors.  
Results in Figure \ref{fig:DecisionBoundaries} are consistent with our experiments in Section \ref{ARD} on more complex datasets.  We see in these decision boundary plots that knowledge distillation from a robust teacher preserves some robustness, while ARD produces a student who closely mimics the teacher.

\section{Related Work}
\label{RelatedWork}

\par 
Early schemes for compressing neural networks involved binarized weights to reduce storage and computation costs \cite{BNN}. Other efforts focused on speeding up calculations via low-rank regularization and pruning weights to reduce computation costs \cite{tai2015convolutional,li2016pruning}. Knowledge distillation teaches a student network to mimic a more powerful teacher \cite{HintonDistillation}.  The student is usually a small, lightweight architecture like MobileNetV2 (MNV2) \cite{Sandler_2018}. 
\par
Knowledge distillation has also been adapted for robustness in a technique called \emph{defensive distillation} \cite{Papernot1}.  In this setting, the teacher and student have identical architectures. An initial network is trained on class labels and then distilled at temperature $t$ onto a network of identical architecture. Defensive distillation improves robustness to a certain $\ell_0$ attack \cite{Papernot1}. However, defensive distillation gains robustness due to gradient masking, and this defense has been broken using $\ell_0$, $\ell_\infty$, and $\ell_2$ attacks \cite{CW1,CW2}.  
\par
Various methods exist that modify networks to achieve robustness.
Some model-specific variants of adversarial training utilize surrogate loss functions to minimize the difference in network output on clean and adversarial data \cite{Zhang2,Miyato}, feature denoising blocks \cite{xie2018feature}, and logit pairing to squeeze logits from clean and adversarial inputs \cite{kannan2018adversarial}. Still other methods, such as JPEG compression, pixel deflection, and image superresolution, are model-agnostic and minimize the effects of adversarial examples by transforming inputs \cite{dziugaite2016study,Prakash_2018,mustafa2019image}.
\par 
Recently, there has been work on defensive-minded compression.  The authors of \cite{compress} and \cite{zhao2018compress} study the preservation of robustness under quantization.  Defensive Quantization involves quantizing a network while minimizing the Lipschitz constant to encourage robustness \cite{lin2019defensive}. While quantization reduces space complexity, it does not reduce the number of Multiply-Add (MAdd) operations needed for inference, although operations performed at lower precision may be faster depending on hardware.  Moreover, Defensive Quantization is not evaluated against strong attackers.  Another compression technique involves pruning \cite{sehwag2019towards}.  Pruning does reduce the number of parameters in a network, but it does not decrease network depth and thus may not accelerate inference.  Additionally, this work does not achieve high compression ratios and does not achieve competitive performance on CIFAR-10.  Pruning maintains the same architectural framework and does not allow a user to compress a state-of-the-art large robust network into a lightweight architecture of their choice.  Creating small robust models is also of interest for the few-shot setting.  Adversarial querying approaches this problem from the meta-learning perspective \cite{goldblum2019adversarially}.


\section{Problem Setup}
\par
Knowledge distillation employs a teacher-student paradigm in which a small student network learns to mimic the output of an often much larger teacher model \cite{HintonDistillation}.  Knowledge distillation in its purist form entails the minimization problem, \begin{align} \min_{\theta}\ell_{\KD}(\theta), \; \; \ell_{\KD}(\theta) = \mathbb{E}_{X\sim \mathcal{D}}\left[\KL(S_\theta^t(X),T^t(X))\right], \end{align} where $\KL$ is KL divergence, $S_\theta$ is a student network with parameters $\theta$, $T$ is a teacher network, $t$ is a temperature constant, and $X$ is an input to the networks drawn from data generating distribution $\mathcal{D}$.  The temperature constant refers to a number by which the logits are divided before being fed into the softmax function.  Intuitively, knowledge distillation involves minimizing the average distance from the student's output to the teacher's output over data from a distribution.  The softmax outputs of the teacher network, also referred to as soft labels, may be more informative than true data labels alone.  In \cite{HintonDistillation}, the authors suggest using a linear combination of the loss function, $\ell_{\KD}(\theta)$, and the cross-entropy between the softmax output of the student network and the one-hot vector representing the true label in order to improve natural accuracy of the student model, especially on difficult datasets.  In this case, we have the loss function \begin{equation}
\begin{aligned}
\ell_{\KD}(\theta) = \mathbb{E}_{(X,d)\sim \mathcal{D}}\big[\alpha t^2\KL(S_\theta^t(X),T^t(X)
\\ +(1-\alpha)\ell(S_{\theta}^t(X),y)\big],
\end{aligned}
\end{equation}
where $\ell$ is the standard cross-entropy loss, and $y$ is the label. In our experiments, we use $\alpha = 1$ except where otherwise noted.  We investigate if students trained using this method inherit their teachers' robustness to adversarial attacks.  We also combine knowledge distillation with adversarial training.

\par
Adversarial training is another method for encouraging robustness to adversarial attacks during training \cite{AdversarialTraining}.  Adversarial training involves the minimax optimization problem, \begin{align} \min_{\theta}\mathbb{E}_{(X,y)\sim\mathcal{D}}\left[ \max_{\|\mathbf{\delta}\|_{p}<\epsilon}L_{\theta}(X+\mathbf{\delta},y) \right],\end{align} where $L_{\theta}(X+\mathbf{\delta},y)$ is the loss of network with parameters $\theta$, input $X$ perturbed by $\mathbf{\delta}$, and label $y$.  Adversarial training encourages a student to produce the correct label in an $\epsilon$-ball surrounding data points.  Virtual Adversarial Training (VAT) and TRADES instead use as a loss function a linear combination of cross-entropy loss and $\KL$ divergence between the network's softmax output from clean input and from adversarial input \cite{Miyato,Zhang2}.  The $\KL$ divergence term acts as a consistency regularizer which trains the  neural network to produce identical output on a natural image and adversarial images generated from the natural image.  As a result, this term encourages the neural network's output to be constant in $\epsilon$-balls surrounding data points.

\par
Knowledge distillation is useful for producing accurate student networks when highly accurate teacher networks exist. However, the resulting student networks may not be robust to adversarial attacks \cite{CW2}.  We combine the central ideas of knowledge distillation and adversarial training to similarly produce robust student networks when robust teacher networks exist.

\par We focus on adversarial robustness to $\ell_{\infty}$ attacks since these are pervasive in the robustness literature.  Thus, we carry out both adversarial training and ARD with FGSM-based PGD $\ell_{\infty}$ attacks similarly to \cite{Madry1,Zhang2}.  We re-implemented the methods from these papers to perform adversarial training and to establish performance baselines.  In our experiments, we use WideResNet (34-10) and ResNet18 teacher models as well as MobileNetV2 (MNV2) students \cite{zagoruyko2016wide,He_2016,Sandler_2018}.  Adversarial training and TRADES teacher models are as described in \cite{Madry1,Zhang2}.

\section{Adversarial robustness is preserved under knowledge distillation}
\label{Preserves}


\par The softmax output of a neural network classifier trained with cross-entropy loss estimates the posterior distribution over class labels on the training data distribution \cite{bridle1990probabilistic}.  Empirical study of distillation suggests that neural networks perform better when trained on the softmax output of a powerful teacher than when trained on only class labels \cite{HintonDistillation}.  The distribution of images generated by adversarial attacks with respect to a particular model and a data generating distribution, $D$, may differ from the distribution of natural images generated by $D$.  Distillation from a naturally trained teacher is known to produce student models which are not robust to adversarial attacks \cite{CW2}.  Thus, we suspected that a non-robust teacher network trained on natural images would be poorly calibrated for estimating posterior probabilities on the distribution of images generated by adversarial attacks.  On the other hand, an adversarially trained teacher network might provide a more accurate estimate of posterior probabilities on this distribution.  We compare the robustness of student models distilled from both naturally trained and adversarially trained teachers.  If we distill from an adversarially trained teacher, will the student inherit robustness?

\par If robustness transfers from teacher to student, we can harness state-of-the-art robust teacher networks to produce accurate, robust, and efficient student networks.  Moreover, since adversarial training is slow due to the bottleneck of crafting adversarial samples for every batch, we could create many different student networks from one teacher, and adversarial training would only need to be performed once.  This routine of training a robust teacher and then distilling onto robust students would be far more time efficient than training many robust student networks individually.  

\subsection{Non-robust teachers produce non-robust students}
\par To establish a baseline for comparison, we distill a non-robust ResNet18 teacher and evaluate against a 20-step PGD attack as in \cite{Madry1}.  We verify known results showing that defensive distillation is ineffective for producing adversarially robust students \cite{CW2}.

\begin{table}[h!]
\centering
\begin{tabular}{|l|l|l|}
\hline
Model                                                   & $\mathcal{A}_{nat}$ & $\mathcal{A}_{adv}$ \\ \hline
ResNet18 teacher                  & 94.75\% & 0.0\% \\ \hline
ResNet18 $\to$ ResNet18                  & 94.92\% & 0.0\% \\ \hline
ResNet18 $\to$ MNV2 & 93.53\% & 0.0\% \\ \hline
\end{tabular}
\caption{Performance of a naturally trained teacher network and its students distilled (with $t=30$) on CIFAR-10, where robust accuracy is with respect to a $20$-step PGD attack as in \cite{Madry1}.  $\mathcal{A}_{nat}$ denotes natural accuracy.}
\label{table:naturallyDistilled}
\end{table}

\subsection{Robust teachers can produce robust students, even distilling on only clean data}
\par Next, we substitute in a robust adversarially trained ResNet18 teacher network and run the same experiments.  We find that our new student networks are far more robust than students of the non-robust teacher (see Table \ref{table:ResNet18ToStudents}).  In fact, the student networks acquire most of the teacher's robust accuracy.  These results confirm that robust lightweight networks may indeed be produced cheaply through knowledge distillation without undergoing adversarial training.

\begin{table}[h!]
\centering
\begin{tabular}{|l|l|l|}
\hline
Student Model                                                   & $\mathcal{A}_{nat}$ & $\mathcal{A}_{adv}$ \\ \hline
AT ResNet18 teacher         & 76.54\% & 44.46\% \\ \hline
AT ResNet18 $\to$ ResNet18      & 76.13\% & 40.13\% \\ \hline
AT ResNet18 $\to$ MNV2 & 76.86\% & 38.21\% \\ \hline
\end{tabular}
\caption{Performance of an adversarially trained ResNet18 teacher network and student networks of various sizes distilled on CIFAR-10, where robust accuracy is with respect to a $20$-step PGD attack as in \cite{Madry1}.}
\label{table:ResNet18ToStudents}
\end{table}

\subsection{Not all robust networks are good teachers, and robustness does not transfer on some datasets}

\par In the previous experiments, we see that a student network may inherit a significant amount of robustness from a robust teacher network during knowledge distillation.  However, some robust teachers are not conducive to this robustness transfer.  We use robust WideResNet (34-10) models trained using adversarial training and TRADES to show that while these models do transfer robustness during knowledge distillation, they transfer less than the weaker ResNet18 teacher network from the previous section (See Table \ref{KDWRN}).  Additionally, a robust WRN teacher model transfers almost no robustness under knowledge distillation against 20-step PGD untargeted attacks on CIFAR-100, a much harder dataset for robustness to untargeted attacks than CIFAR-10 (See Table \ref{KDCIFAR100}).  Further experiments show that robustness transfer diminishes rapidly as we decrease $\alpha$ from our default value of $1$.

\par For these reasons, we develop ARD for distilling a variety of teachers in order to produce robust students.  Using ARD, robustness is preserved on architectures and datasets that do not transfer robustness under vanilla knowledge distillation.

\begin{table}[h!]

\centering
\begin{tabular}{|l|l|l|}
\hline
Model           & $\mathcal{A}_{nat}$ & $\mathcal{A}_{adv}$ \\ \hline
AT WRN teacher & 84.41\% & 45.75\%  \\ \hline
TRADES WRN teacher               & 84.92\% & 56.61\%  \\ \hline
AT WRN $\to$ MNV2& 92.49\% & 5.46\% \\ \hline
TRADES WRN $\to$ MNV2& 85.6\% & 21.69\% \\ \hline
\end{tabular}
\caption{Robust WRN teacher models and their students on CIFAR-10, where robust accuracy is with respect to a $20$-step PGD attack as in \cite{Madry1}.}
\label{KDWRN}

\end{table}
\begin{table}
\centering


\begin{tabular}{|l|l|l|}
\hline
Model           & $\mathcal{A}_{nat}$ & $\mathcal{A}_{adv}$ \\ \hline
AT WRN teacher                & 59.9\% & 28.36\%  \\ \hline
AT WRN $\to$ MNV2   & 25.54\% & 1.30\%  \\ \hline
\shortstack{AT WRN $\to$ MNV2 \\ ($\alpha=0.95$)} & 75.94\% & 0.02\%  \\ \hline
\shortstack{AT WRN $\to$ MNV2 \\ ($\alpha=0.93$)} & 76.38\% & 0.00\%  \\ \hline
\end{tabular}
\caption{Robust teacher network and its students on CIFAR-100, where robust accuracy is with respect to a $20$-step PGD attack as in \cite{Madry1}.}
\label{KDCIFAR100}
\label{table:NotAllTeachers}
\end{table}

\section{Improving the robustness of student models with Adversarially Robust Distillation (ARD)}
\label{ARD}

\par
We combine the central machinery from knowledge distillation, adversarial training, and TRADES/VAT to produce small robust student models from much larger robust teacher models using a method we call \textit{Adversarially Robust Distillation}.  ARD not only produces more robust students than knowledge distillation, but ARD also works for teachers and datasets on which knowledge distillation is ineffective for transferring robustness.  Our procedure is a natural analogue of adversarial training but in a distillation setting.  During standard adversarial training, we encourage a network to produce the ground truth label corresponding to a clean input when the network is exposed to an adversary.  Along the same lines, our method treats the teacher network's softmax output on clean data as the ground truth and trains a student network to reproduce this ground truth when exposed to adversarial examples.  We start with a robust teacher model, $T$, and we train the student model $S_{\theta}$ by solving the following optimization problem ((\ref{eq:ARD}) revisited):
\begin{equation*}
\begin{aligned}
\min_{\theta}\mathbb{E}_{(X,y)\sim \mathcal{D}}\big[\alpha t^2 \KL(S_{\theta}^t(X+\delta_{\theta}),T^t(X))\\ +(1-\alpha)\ell(S_{\theta}^t(X), y)\big],
\end{aligned}
\end{equation*}
where $\delta_\theta = \argmax_{\|\delta\|_{p}<\epsilon}\ell(S_{\theta}^t(X+\delta), y)$, $\ell$ is cross-entropy loss, and we divide the logits of both student and teacher models by temperature term $t$ during training.  The $t^2$ term is used as in \cite{HintonDistillation} since dividing the logits shrinks the gradient.  The cross-entropy loss, which encourages natural accuracy, is a standard training loss.  Thus, $\alpha$ is a hyperparameter that prioritizes similarity to the teacher over natural accuracy.  In our experiments, we set $\alpha=1$ except where otherwise noted, eliminating the cross-entropy term.  We find that lower values of $\alpha$ are useful for improving performance on harder classification tasks.  Our training routine involves the following procedure:

\SetArgSty{textnormal}
\begin{algorithm}[h!]
$\text{\bf{Require:}}$ Student and teacher networks $S$ and $T$, learning rate $\gamma$, dataset $\{(\mathbf{x}_i, y_i)\}$, number of steps, K, per PGD attack, and $\epsilon$ maximum attack radius.\\
Initialize $\theta$, the weights $S$\;
 \For{$\text{Epoch}$ = 1,...,$N_{epochs}$}{
  \For{$\text{Batch}$ = 1,...,$N_{batches}$}{
   Construct adv. example $\mathbf{x}'_i$ for each $\mathbf{x}_i\in$ Batch by maximizing cross-entropy between $S_{\theta}(\mathbf{x}_i')$ and $\mathbf{y}_i$ constrained to $\|\mathbf{x}_i-\mathbf{x}_i'\|_p<\epsilon$ using K-step PGD.\\
   Compute $\nabla_{\theta} \ell_{\ARD}(\{\mathbf{x}_i\}, \theta) = \sum_{i}\nabla_{\theta}[\alpha t^2 \KL(S_{\theta}^t(\mathbf{x}_i'),T^t(\mathbf{x}_i))$ $+(1-\alpha)\ell(S_{\theta}^t(X), \mathbf{y})]$, over the current batch.\\
   $\theta \leftarrow \theta - \gamma \nabla_{\theta} \ell_{\ARD}(\{\mathbf{x}_i\}, \theta)$
  }
 }
\caption{Adversarially Robust Distillation (ARD)}
\end{algorithm}

\subsection{ARD works with teachers that fail to transfer robustness under knowledge distillation}
\par In Section \ref{Preserves}, we saw that that some teacher networks do not readily transfer robustness.  We see through our experiments in Table \ref{table:ResultsTableCIFAR-10} that ARD is able to create robust students from teachers whose robustness failed to transfer during knowledge distillation.  TRADES and adversarially trained MobileNetV2 (MNV2) models are each outperformed in both natural and robust accuracy simultaneously by an ARD variant.  In these experiments, the WideResnet teacher model contains $20\times$ as many parameters and performs $70\times$ as many MAdd operations as the MobileNetV2 student.

\begin{table}[h!]
\centering
\begin{tabular}{|l|l|l|}
\hline
Model                                                    & $\mathcal{A}_{nat}$ & $\mathcal{A}_{adv}$ \\ \hline
TRADES WRN teacher                                & 84.92\% & 56.61\% \\ \hline
AT MNV2                        & 80.50\% & 46.90\% \\ \hline
TRADES MNV2                          & 83.59\% & 44.79\% \\ \hline
TRADES WRN $\xrightarrow{\text{\textbf{ARD}}}$ MNV2                  & 82.63\% & \textbf{50.42\%} \\ \hline
\shortstack{TRADES WRN $\xrightarrow{\text{\textbf{ARD}}}$ MNV2 \\ ($\alpha = 0.95$)}& \textbf{84.70\%} & 46.28\% \\ \hline
\end{tabular}
\caption{Performance on CIFAR-10, where robust accuracy is w.r.t. a $20$-step PGD attack as in \cite{Madry1}.}
\label{table:ResultsTableCIFAR-10}
\end{table}

\subsection{ARD works on datasets where knowledge distillation fails}
\par In Section \ref{Preserves}, we saw that a student network inherited little robustness from an adversarially trained teacher network on CIFAR-100.  This dataset is very difficult to protect from untargeted attacks because it contains many classes that are similar in appearance.  In Table \ref{table:ResultsTableCIFAR-100}, we see that a MobileNetV2 student model trained on CIFAR-100 using ARD from an adversarially trained WideResNet is significantly more robust than an adversarially trained MobileNetV2.  In fact, the ARD model is nearly as robust as its teacher.


\begin{table}[h!]
\centering
\begin{tabular}{|l|l|l|}
\hline
Model                                                    & $\mathcal{A}_{nat}$ & $\mathcal{A}_{adv}$ \\ \hline
AT WRN teacher                              & 59.90\% & 28.36\% \\ \hline
AT MNV2                        & 55.62\% & 22.80\% \\ \hline
AT WRN $\xrightarrow{\text{\textbf{ARD}}}$ MNV2 ($\alpha = 0.93$)& 55.47\% & \textbf{27.64}\% \\ \hline
\end{tabular}
\caption{Performance on CIFAR-100, where robust accuracy is w.r.t. a $20$-step PGD attack as in \cite{Madry1}.}
\label{table:ResultsTableCIFAR-100}
\end{table}

\subsection{ARD can produce networks more robust than their teacher}
\par In some experiments, ARD student networks are more robust than their teacher.  Interestingly, this behavior does not depend on distilling from a high-capacity network to a low capacity network; distilling Resnet18 onto itself and MobileNetV2 onto itself using ARD results in far better robustness than adversarial training alone.

We seek to understand if this increased robustness is caused by differences between the student and teacher architectures, and so we use ARD to distill a teacher network onto a student network with an identical architecture.  In our experiments, ARD boosted the robustness of both the ResNet18 and MobileNetV2 models (See Table \ref{table:ResNet18Teacher}).

\begin{table}[h]
\centering
\begin{tabular}{|l|l|l|}
\hline
Model                                                    & $\mathcal{A}_{nat}$ & $\mathcal{A}_{adv}$ \\ \hline
AT ResNet18 & 76.54\% & 44.46\% \\ \hline
AT MNV2                & 80.50\% & 46.90\% \\ \hline
AT ResNet18 $\xrightarrow{\text{\textbf{ARD}}}$ ResNet18 & 79.49\% & \textbf{51.21}\% \\ \hline
AT MNV2 $\xrightarrow{\text{\textbf{ARD}}}$ MNV2 & 81.22\% & 47.95\% \\ \hline
AT ResNet18 $\xrightarrow{\text{\textbf{ARD}}}$ MNV2 & 79.47\% & 50.22\% \\ \hline

\end{tabular}
\caption{Performance on CIFAR-10, where robust accuracy is w.r.t. a $20$-step PGD attack as in \cite{Madry1}.}
\label{table:ResNet18Teacher}
\end{table}

\subsection{Accelerating ARD using fast adversarial training methods}
\label{freetraining}
\par The ARD procedure described above takes approximately the same amount of time as adversarial training.  Adversarial training is slow since it requires far more gradient calculations than natural training.  Several methods have been proposed recently for accelerating adversarial training \cite{freetraining,propagateonce}.  We similarly accelerate performance for ARD by adapting ``free'' adversarial training to distillation.  This version, \emph{Fast-ARD}, described in Algorithm \ref{alg:Fast-ARD}, is equally fast to knowledge distillation (see Table \ref{table:FreeTrainingTimes} for a list of training times).  During training, we replay each mini-batch several times in a row.  On each replay, we simultaneously compute the gradient of the loss w.r.t. the image and parameters using the same backward pass.  Then, we update the adversarial attack and the network's parameters simultaneously.  Empirically, Fast-ARD produces less robust students than the full ARD above, but it produces higher robust accuracy compared to models with identical architectured trained using existing accelerated free adversarial training methods as seen in Table \ref{table:FreeTrainingResults}.  Furthermore, Fast-ARD from a TRADES WideResNet onto MobileNetV2 produces a more robust student than our most robust MobileNetV2 produced during vanilla knowledge distillation and in the same amount of training time.  Our accelerated algorithm is detailed in Algorithm \ref{alg:Fast-ARD}.

\SetArgSty{textnormal}
\begin{algorithm}[h!]
$\text{\bf{Requires:}}$ Student and teacher networks $S$ and $T$, learning rate $\gamma$, norm $p$, dataset $\{(\mathbf{x}_i,y_i)\}$, and attack step-size $r$ and radius $\epsilon$\\
Initialize $\theta$, the weights of network $S$, and set $\bm{\delta}=0$.\\
 \For{$\text{Epoch}$ = 1,...,$\frac{N_{epochs}}{m}$}{
  \For{$\text{Batch}$ = 1,...,$N_{batches}$}{
   \For{j = 1,..., m}{
   For $\mathbf{x}_i\in$ Batch, find new perturbation $\delta'_i$ by maximizing cross-entropy between $S_{\theta}(\mathbf{x}_i+\delta_i+\delta'_i)$ and $\mathbf{y}_i$ over $\delta'_i$, constrained to $\|\delta_i+\delta'_i\|_p<\epsilon$, using a 1-step PGD attack.\\
   Compute the gradient of the loss function $\nabla_{\theta} \ell_{\ARD}(\{\mathbf{x}_i\}, \theta) = \sum_{i}\nabla_{\theta}[\alpha t^2 \KL(S_{\theta}^t(\mathbf{x}_i'),T^t(\mathbf{x}_i))+(1-\alpha)\ell(S_{\theta}^t(X), \mathbf{y})]$, over current batch.\\
   $\delta_i \leftarrow \delta_i + \delta'_i$\\
   $\theta \leftarrow \theta - \gamma \nabla_{\theta} \ell_{\ARD}(\{\mathbf{x}_i\}, \theta)$
   }
  }
 }
 \caption{Fast-ARD with free adversarial training}
 \label{alg:Fast-ARD}
\end{algorithm}

\begin{table}[h!]
\centering
\begin{tabular}{|l|l|l|}
\hline
Model                                                    & $\mathcal{A}_{nat}$ & $\mathcal{A}_{adv}$ \\ \hline
\shortstack{Free trained MNV2 (m=4)}                   & 82.63\% & 23.13\% \\ \hline
\shortstack{TRADES WRN $\xrightarrow{\text{\textbf{F-ARD}}}$ MNV2\\ (m=4)}                   & \textbf{83.51}\% & \textbf{37.07}\%  \\ \hline
\shortstack{Free trained MNV2 (m=8)}                   & 72.30\% & 27.96\% \\ \hline
\shortstack{TRADES WRN $\xrightarrow{\text{\textbf{F-ARD}}}$ MNV2\\ (m=8)}                   & 76.38\% & 36.85\%  \\ \hline
\end{tabular}
\caption{Performance of MobileNetV2 classifiers, free trained and Fast-ARD, on CIFAR-10, where robust accuracy is with respect to a $20$-step PGD attack as in \cite{Madry1}.  ``$\xrightarrow{\text{\textbf{F-ARD}}}$'' denotes ``Fast-ARD onto''.}
\label{table:FreeTrainingResults}
\end{table}

\begin{table}[h!]
\centering
\begin{tabular}{|l|l|}
\hline
Model & Time (hrs)   \\ \hline
AT MNV2 & 41.09  \\ \hline
TRADES WRN $\rightarrow$ MNV2 & 5.13   \\ \hline
TRADES WRN $\xrightarrow{\text{\textbf{ARD}}}$ MNV2 & 41.06   \\ \hline
TRADES WRN $\xrightarrow{\text{\textbf{F-ARD}}}$ MNV2 (m=4) & 5.15   \\ \hline
TRADES WRN $\xrightarrow{\text{\textbf{F-ARD}}}$ MNV2 (m=8) & $\textbf{5.10}$   \\ \hline
\end{tabular}
\caption{Training times for adversarial training, clean distillation, ARD, and Fast-ARD. Each model was trained with 200 parameter updates for each training image (equivalent of 200 epochs). All models were trained on CIFAR-10 with a single RTX 2080 Ti GPU and identical batch sizes. The AT model and the ARD model were trained with a $10$-step PGD attack.  ``$\xrightarrow{\text{\textbf{F-ARD}}}$'' denotes ``Fast-ARD onto''.}
\label{table:FreeTrainingTimes}
\end{table}

\begin{table*}[h!]
\centering
\begin{tabular}{|l|l|l|l|l|l|}
\hline
Model  & MI-FGSM$_{20}$ & DeepFool & 1000-PGD & 20-PGD 100-restarts \\ \hline
AT MNV2& 50.82\%  & 57.74\% & 46.51\% & 46.79\% \\ \hline
$\xrightarrow{\text{\textbf{ARD}}}$ MNV2& \textbf{55.16}\% &  \textbf{64.61}\% & \textbf{49.98}\% & \textbf{50.30}\% \\ \hline
Free trained MNV2 (m=4) & 30.60\% & 41.09\% & 22.23\% & 22.94\% \\ \hline
$\xrightarrow{\text{\textbf{F-ARD}}}$ MNV2(m=4) & \textbf{44.78}\% &  \textbf{60.03}\% & \textbf{36.01}\% & \textbf{36.88}\% \\ \hline
\end{tabular}
\caption{Robust validation accuracy of adversarially trained and free trained MobileNetV2 and TRADES WRN ARD (and Fast-ARD) onto MobileNetV2 on CIFAR-10 under various attacks.  All attacks use $\epsilon = \frac{8}{255}$.}
\label{table:AttackExperimentsCIFAR10}
\end{table*}

\subsection{ARD and Fast-ARD models are more robust than their adversarially trained counterparts}
While 20-step PGD is a powerful attack, we also test ARD against other $\ell_\infty$ attackers including Momentum Iterative Fast Gradient Sign Method \cite{dong2018boosting}, DeepFool \cite{DeepFool}, 1000-step PGD, and PGD with random restarts.  We find that ARD and Fast-ARD outperform adversarial training and free training respectively across all attacks we tried (see Table \ref{table:AttackExperimentsCIFAR10}).

\section{Space and time efficiency of student and teacher models}
\label{models}
\par We perform our experiments for ARD with WideResNet (34-10) and ResNet18 teacher models as well as a MobileNetV2 student model.  We consider two network qualities for quantifying compression.  First, we study space efficiency by counting the number of parameters in a network.  Second, we study time complexity.  To this end, we compute the multiply-add (MAdd) operations performed during a single inference.  The real time elapsed during this inference will vary as a result of implementation and deep learning framework, so we use MAdd, which is invariant under implementation and framework, to study time complexity.
\par The WideResNet and ResNet18 teachers we employ contain $\sim 46.2$M and $\sim 11.2$M parameters respectively, while the MobileNetV2 student contains $\sim 2.3$M parameters.  A forward pass through the WRN and ResNet18 teacher models takes $\sim 13.3$B MAdd operations and $\sim 1.1$B MAdd operations respectively, while a forward pass through the student model takes $\sim 187$M MAdd operations.  To summarize these network traits, compared to a WideResnet (34-10) teacher, the MobileNetV2 student model:
\begin{itemize}
    \item Contains $\sim 5\%$ as many parameters
    \item Performs $\sim 1.4\%$ as many MAdd operations during a forward pass
\end{itemize}

\section{Discussion}
\label{Discussion}

\par We find that knowledge distillation allows a student network to absorb a large amount of a teacher network's robustness to adversarial attacks, even when the student is only trained on clean data.  However, in some cases, a distilled student model is still far less robust than the teacher.  To improve student robustness, we introduce Adversarially Robust Distillation (ARD).  In our experiments, student models trained using our method outperform similar networks trained using adversarial training in robust and often natural accuracy.  Our models exceed state-of-the-art performance on CIFAR-10 and CIFAR-100 benchmarks.  Furthermore, we develop a free adversarial training variant of ARD and demonstrate appreciably accelerated performance.
\par Recent work on distillation has produced significant improvements over vanilla knowledge distillation \cite{KDFM}.  We believe that Knowledge Distillation with Feature Maps could improve both natural and robust accuracy of student networks.  Adaptive data augmentation like AutoAugment \cite{Autoaugment} may also improve performance of both normal knowledge distillation for robust teachers and ARD.  Finally, with the recent publication of fast adversarial training methods \cite{freetraining,propagateonce}, we hope to further accelerate ARD.

\section{Experimental details}
We train our models for 200 epochs with SGD, momentum of $0.9$, and weight-decay of $2e-4$.  Fast-ARD models are trained for $\frac{200}{m}$ epochs so that they take the same amount of time as natural distillation.  We use an initial learning rate of 0.1, and we decrease the learning rate by a factor of 10 on epochs 100 and 150 (epochs $\frac{100}{m}$ and $\frac{150}{m}$ for Fast-ARD).  We use a temperature term of $30$ for CIFAR-10 and $5$ for CIFAR-100.  To craft adversarial examples during training, we use FGSM-based PGD with 10 steps, $\ell_\infty$ attack radius of $\epsilon=\frac{8}{255}$, a step size of $\frac{2}{255}$, and a random start.

\par A PyTorch implementation of ARD can be found at: \\ https://github.com/goldblum/AdversariallyRobustDistillation

\section*{Acknowledgments}
\par This research was generously supported by DARPA, including the GARD program, QED for RML, and the Young Faculty Award program.  Further funding was provided by the AFOSR MURI program, and the National Science Foundation (DMS).

\bibliographystyle{aaai}
\bibliography{refs.bib}

\section*{Appendix}

\subsection*{Appendix A: The effects of temperature, $\alpha$, and data augmentation for knowledge distillation of robust teacher models}

\par We show the results of experiments in Table \ref{table:Non_ARD_Temperature} and Table \ref{table:Non_ARD_alpha}, varying temperature and $\alpha$.  We see that only very low temperature terms are not conducive to robustness preservation, but knowledge distillation is not highly sensitive to temperature, and a wide array of temperature terms are effective for producing robust students.  On the other hand, robustness transfer decays rapidly when $\alpha$ decreases.  This effect produces a robustness-accuracy tradeoff.
\par Data augmentation yields dramatic improvements for robustness transfer as the student gets to see the teacher's behavior at more data points, as can be seen in Table \ref{table:DataAugmentation}.  A natural idea is to teach the student the teacher's behavior at adversarial points as well.  However, this data augmentation technique greatly decreases training speed and does not provide significant improvement.

\begin{table}[h!]
\centering
\begin{tabular}{|l|l|l|}
\hline
Temperature                                                   & $\mathcal{A}_{nat}$ & $\mathcal{A}_{adv}$ \\ \hline
1  & 77.37\% & 35.44\%  \\ \hline
10 & 78.1\% & 38.55\% \\ \hline
30 & 76.86\% & 38.21\% \\ \hline
50 & 76.28\% & 38.5\% \\ \hline
100 & 76.44\% & 38.65\% \\ \hline
\end{tabular}
\caption{Performance of adversarially trained ResNet18 teacher network distilled onto MobileNetV2 (MN) with different temperature terms on CIFAR-10, where robust accuracy is with respect to a $20$-step PGD attack as in [15].}
\label{table:Non_ARD_Temperature}

\end{table}
\begin{table}[h!]

\centering
\begin{tabular}{|l|l|l|}
\hline
$\alpha$                                                   & $\mathcal{A}_{nat}$ & $\mathcal{A}_{adv}$ \\ \hline
1.00  & 76.86\% & 38.21\%  \\ \hline
0.99 & 82.93\% & 28.36\% \\ \hline
0.95 & 91.58\% & 9.65\% \\ \hline
0.90 & 92.34\% & 7.96\% \\ \hline
0.70 & 92.33\% & 7.54\% \\ \hline
0.50 & 92.14\% & 2.73\% \\ \hline
\end{tabular}
\caption{Performance of adversarially trained ResNet18 teacher network distilled onto MobileNetV2 (MN) with different $\alpha$ terms on CIFAR-10, where robust accuracy is with respect to a $20$-step PGD attack as in [15].}
\label{table:Non_ARD_alpha}
\end{table}

\begin{table}[h!]
\centering
\scalebox{0.87}{
\begin{tabular}{|l|l|l|}
\hline
Data Augmentation                                                   & $\mathcal{A}_{nat}$ & $\mathcal{A}_{adv}$ \\ \hline
No augmentation  & 75.76\% & 32.05\%  \\ \hline
Horizontal flips  & 76.24\% & 35.39\% \\ \hline
Random crops & 76.31\% & 37.31\% \\ \hline
Both flips and crops & 76.86\% & 38.21\% \\ \hline
Adv. examples generated from teacher & 76.76\% & 38.33\% \\ \hline
\end{tabular}}
\vspace{1.5 mm}
\caption{Performance of adversarially trained WideResNet distilled onto MobileNetV2 (MN) with various data augmentation on CIFAR-10, where robust accuracy is with respect to a $20$-step PGD attack as in [15].}
\label{table:DataAugmentation}
\end{table}



\subsection*{Appendix B: Notes on loss functions for ARD}
\par We tried other versions of the loss function with the addition of a (non-adversarial) knowledge distillation $\KL$ divergence term, which increases natural accuracy slightly but sharply decreases robust accuracy, and the addition of cross-entropy between the softmax of the adversarially attacked student and the one-hot label vectors, which is redundant in combination with the ARD loss and empirically harms robustness.  We also explored using $T^t(\mathbf{x}_i')$ instead of $T^t(\mathbf{x}_i)$ in the $\KL$ divergence term in the loss function, but the accuracy, both natural and robust, decreased.  Additionally, passing an adversarial example through the teacher model will slow down training and sharply increase memory consumption if the teacher is much larger than the student.  Our method does not require forward passes through the teacher during training as long as logits of the training data are stored ahead of time.  We explored generating adversarial attacks during training by maximizing $\KL$ divergence instead of cross-entropy.  This technique lowers natural accuracy without improving robust accuracy.
\par In our training routine, we employ data augmentation in the form of random crops and horizontal flips.  We intend to explore adaptive data augmentation methods tailored specifically for robust distillation.

\subsection*{Appendix C: ARD with naturally trained teacher models}
\par ARD encourages a student to produce, for all images within an $\epsilon$-ball of a data point, the teacher's output at that data point.  Thus, it seems reasonable to try using ARD with a naturally trained teacher model.  As evident in Table \ref{table:NaturalARD}, naturally trained teachers produce robust students, but these students may be less robust than those of robust teachers and less robust than adversarially trained models with identical architecture.

\begin{table}[h!]
\centering
\begin{tabular}{|l|l|l|}
\hline
Model                                                    & $\mathcal{A}_{nat}$ & $\mathcal{A}_{adv}$ \\ \hline
ResNet18$\xrightarrow{\text{\textbf{ARD}}}$ MobileNetV2              & 84.18\% & 44.61\% \\ \hline
WRN $\xrightarrow{\text{\textbf{ARD}}}$ MobileNetV2              & 84.43\% & 42.51\% \\ \hline
\end{tabular}
\vspace{1.5 mm}
\caption{Performance of MobileNetV2 students distilled from naturally trained teachers on CIFAR-10, where robust accuracy is with respect to a $20$-step PGD attack as in [15].}
\label{table:NaturalARD}
\end{table}

\subsection*{Appendix D: Improving the speed of ARD by reducing the number of attack steps}
\par Another way to improve the speed of ARD training is to reduce the number of attack steps in order to reduce the number of gradient calculations.  In our experiments shown in Table \ref{table:AttackSteps}, reducing the number of attack steps improves natural accuracy while decreasing robust accuracy.  This procedure has a similar effect to reducing $\alpha$.  We suggest this strategy over reducing $\alpha$ since it has the added benefit of accelerating training.

\begin{table}[h!]
\label{Steps}
\centering
\begin{tabular}{|l|l|l|}
\hline
Attack Steps (training)                                                  & $\mathcal{A}_{nat}$ & $\mathcal{A}_{adv}$ \\ \hline
4    & 84.28\% & 46.49\%  \\ \hline
6    & 83.38\% & 48.35\% \\ \hline
10   & 82.63\% & 50.42\% \\ \hline
\end{tabular}
\vspace{1.5 mm}
\caption{Performance of TRADES WideResNet distilled onto MobileNetV2 using ARD with adversaries generated using different numbers of attack steps on CIFAR-10. Robust accuracy is with respect to a $20$-step PGD attack as in [15].}
\label{table:AttackSteps}
\end{table}

\subsection*{Appendix E: Sensitivity of ARD to temperature and $\alpha$}
\par Compared to knowledge distillation for preserving robustness, ARD is less sensitive to the temperature parameter.  We found that varying the temperature parameter does not significantly impact either the natural or robust accuracy of the resulting student (See Table \ref{table:ARDTemperature}).  Like with knowledge distillation, varying $\alpha$ presents an accuracy-robustness tradeoff (See Table \ref{table:ARDAlpha}).  However, unlike with knowledge distillation, we find that under ARD, robust accuracy decays far slower as $\alpha$ decreases so that ARD is less sensitive to this parameter.

\begin{table}[h!]
\centering

\begin{tabular}{|l|l|l|}
\hline
Temperature                                                   & $\mathcal{A}_{nat}$ & $\mathcal{A}_{adv}$ \\ \hline
1  & 81.41\% & 49.57\%  \\ \hline
5  & 82.24\% & 48.77\% \\ \hline
10 & 82.33\% & 48.93\% \\ \hline
30 & 82.63\% & 50.42\% \\ \hline
50 & 82.14\% & 48.97\% \\ \hline
\end{tabular}
\caption{Performance of TRADES WideResNet distilled onto MobileNetV2 using ARD with different temperature terms on CIFAR-10, where robust accuracy is with respect to a $20$-step PGD attack as in [15].}
\label{table:ARDTemperature}
\end{table}
\begin{table}[h!]

\centering
\begin{tabular}{|l|l|l|}
\hline
$\alpha$                                                   & $\mathcal{A}_{nat}$ & $\mathcal{A}_{adv}$ \\ \hline
1.00         & 82.63\% & 50.42\%  \\ \hline
0.95        & 84.7\% & 46.28\% \\ \hline
0.90 & 86.58\% & 41.16\% \\ \hline
0.70 & 90.57\% & 25.18\% \\ \hline
0.50 & 83.00\% & 13.09\% \\ \hline
\end{tabular}
\caption{Performance of TRADES WideResNet distilled onto MobileNetV2 using ARD with different values of $\alpha$ on CIFAR-10, where robust accuracy is with respect to a $20$-step PGD attack as in [15].}
\label{table:ARDAlpha}
\end{table}

\end{document}